\documentclass[letterpaper]{article} % DO NOT CHANGE THIS
\usepackage{aaai25}  % DO NOT CHANGE THIS
\usepackage{times}  % DO NOT CHANGE THIS
\usepackage{helvet}  % DO NOT CHANGE THIS
\usepackage{courier}  % DO NOT CHANGE THIS
\usepackage[hyphens]{url}  % DO NOT CHANGE THIS
\usepackage{graphicx} % DO NOT CHANGE THIS
\usepackage{natbib}  % DO NOT CHANGE THIS AND DO NOT ADD ANY OPTIONS TO IT
\usepackage{caption} % DO NOT CHANGE THIS AND DO NOT ADD ANY OPTIONS TO IT
\usepackage{algorithm}
\usepackage{algorithmic}
\usepackage{amsmath} 
\usepackage{amssymb}
\usepackage{color}
\usepackage{makecell}

\usepackage{url}
\usepackage{newfloat}
\usepackage{listings}
\DeclareCaptionStyle{ruled}{labelfont=normalfont,labelsep=colon,strut=off} % DO NOT CHANGE THIS
\floatstyle{ruled}
\newfloat{listing}{tb}{lst}{}
\floatname{listing}{Listing}
\title{CustomTTT: Motion and Appearance Customized Video Generation via Test-Time Training}
\author{
    Xiuli Bi\textsuperscript{\rm 1},
    Jian Lu\textsuperscript{\rm 1},
    Bo Liu\textsuperscript{\rm 1},
    Xiaodong Cun\textsuperscript{\rm 2}\thanks{Corresponding Author},
    Yong Zhang\textsuperscript{\rm 3},
    Weisheng Li\textsuperscript{\rm 1},
    Bin Xiao\textsuperscript{\rm 1,\rm 4}
}

\affiliations{
    \textsuperscript{\rm 1}~School of Computer Science and Technology, Chongqing University of Posts and Telecommunications, Chongqing, China,\\
    \textsuperscript{\rm 2}~GVC Lab, Great Bay University, Dongguan, China,
    \textsuperscript{\rm 3}~Meituan,
    \textsuperscript{\rm 4}~Jinan Inspur Data Technology Co., Ltd., Jinan, China\\
    \{boliu, bixl, liws, xiaobin\}@cqupt.edu.cn,
    s230201078@stu.cqupt.edu.cn,
    cun@gbu.edu.cn,
    zhangyong201303@gmail.com

}

\usepackage{xspace}
\usepackage{booktabs}
\usepackage{multirow}
\usepackage[table]{xcolor}

\makeatletter
\DeclareRobustCommand\onedot{\futurelet\@let@token\@onedot}
\def\@onedot{\ifx\@let@token.\else.\null\fi\xspace}

\def\eg{\emph{e.g}\onedot} 
\def\ie{\emph{i.e}\onedot} 
 
\def\etc{\emph{etc}\onedot}

\makeatother
\usepackage{bibentry}
\begin{document}

\twocolumn[{
\renewcommand\twocolumn[1][]{#1}%

\maketitle

\begin{center}
    % \captionsetup{type=figure}
    % \vspace{-4.5\baselineskip}
    \includegraphics[width=1.0\textwidth]{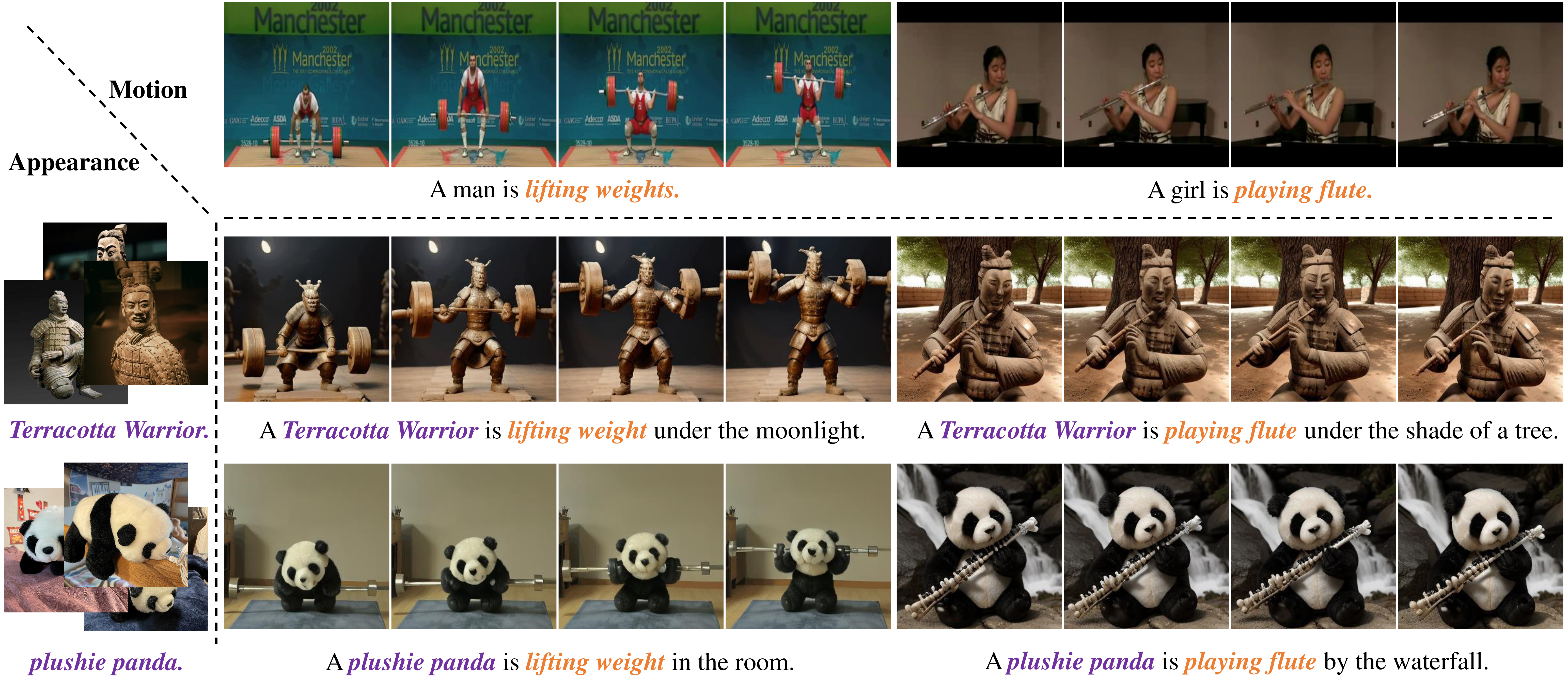}
    \vspace{-2em}
    \captionof{figure}{
    % \textit{ \xiaodong{A teaser Figure to show the effectiveness of our method}}
    Given a \textit{single} video for motion reference and \textit{a few} images for appearance reference, our method can generate customized videos with multiple customized concepts in terms of the combinations of appearance and motion.
    }
\end{center}
}]

\renewcommand{\thefootnote}{\fnsymbol{footnote}}
\footnotetext[1]{~Corresponding Author.}

\let\thefootnote\relax\footnotetext{Copyright \copyright\space 2025,
Association for the Advancement of Artificial Intelligence (www.aaai.org).
All rights reserved.}

\begin{abstract}
Benefiting from large-scale pre-training of text-video pairs, current text-to-video~(T2V) diffusion models can generate high-quality videos from the text description. Besides, given some reference images or videos, the parameter-efficient fine-tuning method, \ie LoRA, can generate high-quality customized concepts, \eg, the specific subjects or the motions. However, combining the trained multiple concepts from different references into a single network shows obvious artifacts. To this end, we propose CustomTTT, where we can joint custom the appearance and the motion of the given video easily. In detail, we first analyze the prompt influence in the current video diffusion model and find the LoRAs are only needed for the specific layers for appearance and motion customization. Besides, since each LoRA is trained individually, we propose a novel test-time training technique to update parameters after combination utilizing the trained customized models. We conduct detailed experiments to verify the effectiveness of the proposed methods. Our method outperforms several state-of-the-art works in both qualitative and quantitative evaluations. 
\end{abstract}
\begin{links}
    \link{Code}{https://github.com/RongPiKing/CustomTTT}
\end{links}

\section{Introduction}

In the realm of text-to-video~(T2V) diffusion models~\cite{Animatediff, videocrafter1, videocrafter2, Zeroscope}, significant progress has been achieved through large-scale pre-training of text-video pairs, enabling the generation of high-quality videos from text descriptions. However, these pre-trained models present difficulties in generating specific objects or motions that are hard to describe from text prompts only. 

To solve this problem, early model customization methods, \eg, Dreambooth~\cite{Dreambooth}, LoRA~\cite{LoRA}, are proposed to produce customized concepts, such as specific subjects or motions, when providing reference images or videos. However, these methods focus only on the specific \textit{one} concept. There are also some works for the multi-concept generation. However, they are mostly works for the two different subjects~\cite{celeb-basis, customDiffusion}. For the customized appearance and motion, the pioneer works~\cite{Dreamvideo} still face the problem of appearance overfitting and low-fidelity. 

We aim for multiple customization methods including the subject and the motion.
Inspired by the current LoRA merged methods~\cite{yadav2024ties,yu2024language}, we can train different LoRAs and merge them for this purpose. Differently, in our task, we can naturally select different layers~(\ie, spatial and temporal layers) in video diffusion models for motion and appearance customization. However, the resulting video shows heavy artifacts. Thus, we should fine-tune the model when merging two LoRAs to reduce the conflicts during the merging process further.

Therefore, we present CustomTTT for joint motion and appearance customization. In detail, to find the most criteria layers for appearance and motion modeling in the VDM, we perform the text-embedding replacement experiment inspired by the text-to-image customization method~\cite{B-LoRA}. On the other hand, after training and interpolating two LoRA adapters for motion and appearance customization, we utilize the pre-trained single LoRA model to produce a suitable reference latent for distillation in a certain step. Thus, the multi-customized model can learn to guarantee the motion and appearance from their single customized reference. We conduct the experiments over various motions and the subject reference. Compared with state-of-the-art methods, our method shows much better visual quality and text-video alignment under the multi-customization settings.

Overall, the contribution can be summarized as:

\begin{itemize}
    \item We analyze the prompt embedding importance in the T2V model for the first time and train LoRA adapters on the specific layers for better multiple-concept customization.
    \item We design a novel test-time training process to improve the text-video alignment in multiple concept customization.
    \item The experiments show the advantage of the proposed method over current methods.
\end{itemize}

\section{Related Work}

\subsection{Text-to-Video Diffusion Model}
The Diffusion Model~\citep{ho2020denoising, ddim} has reinvented the research in video generation with the pre-trained text-to-image latent diffusion model~\citep{LDM} as the basis. 
Different from previous methods which only can generate the video in the specific domain~(\eg, face~\citep{fomm, sadtalker}, body~\citep{mraa}), these text-to-video diffusion models~\cite{Animatediff, videocrafter1, videocrafter2, Zeroscope} are based on the pre-trained text-to-image model~\cite{LDM} and train the temporal layers to model motion. After large-scale training, these methods can generate open-domain realistic videos from text prompts. However, the specific object or motion is hard to describe via text only. Thus, training the diffusion-based model for the specific appearance or motion is critical.

\subsection{Model Customization} Fine-tuning the pre-trained large text-to-image/video models~\cite{LDM, videocrafter1, videocrafter2} to generate the customized image~(or video) is naturally. Since the specific object/motion can not be described easily in text. Early works~\cite{Dreambooth, LoRA, Textual-Inversion} focus on single subject customization via additional training, as well as multiple subjects~\cite{celeb-basis, customDiffusion} and also for video~\cite{jiang2024videobooth}. Recent works are to customize the motion of the video by decoupling the training of the spatial and temporal~\cite{Motiondirector, wang2024motion, ren2024customize}. The most related work to our method is DreamVideo~\cite{Dreamvideo}, they train additional adapters for motion and appearance customization. However, the performance is still restricted by the appearance overfit or the motion correctness.

\subsection{Test-time Training}
Test-time training~(TTT) aims to improve the performance of the specific sample when testing without the label. Previous works have proved its efficiency on image classification~\cite{wang2020tent, NEURIPS2022_bcdec1c2,pmlr-v119-sun20b,NEURIPS2020_85690f81}, video object segmention~\cite{dattt,wang2023test}. The core of the TTT is to design a self-supervised proxy function, \eg, the rotation~\cite{pmlr-v119-sun20b}, the masked image reconstruction~\cite{NEURIPS2022_bcdec1c2}, so as the pre-trained knowledge can be updated via this proxy gradient. In our multiple customizations of appearance and motion, we use this method in diffusion-based customization to update the parameters after the LoRA combination. We construct the relationship between the multiple and the single customization model.

\begin{figure*}[!t]
\centering
\includegraphics[width=\textwidth, height=0.6\textwidth]{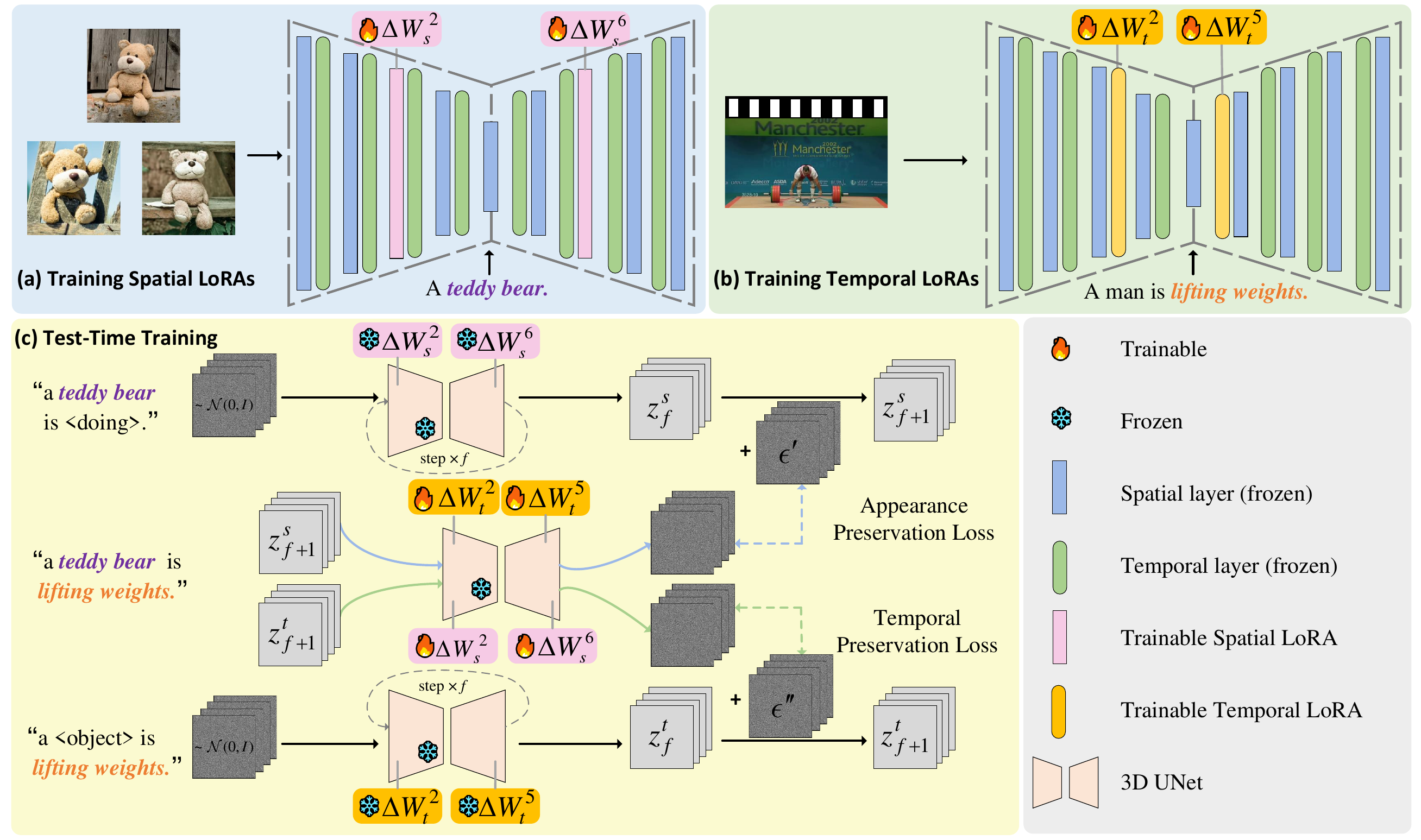}
\vspace{-2em}
\caption{
The overall pipeline. We first train the LoRAs on the specific layers for appearance~(a) and motion~(b) customization individually. Then, we design a test-time training method to further improve the results when combining.}
\vspace{-1em}
\label{fig:overall}
\end{figure*}

\section{Method}

We aim to generate a highly customized video from text, where the motion and appearance are from \textit{different} given references. These references may include a video for motion guidance and 3-5 images for appearance guidance.
To achieve this goal, we utilize the pre-trained Text-to-Video Diffusion Models~\cite{Animatediff} as the base model and train the LoRAs~\cite{LoRA} for motion and appearance customization, respectively.
However, directly incorporating individually trained temporal and spatial LoRAs into a single pre-trained model for multiple customizations will typically fail. To this end, we first perform an in-depth prompt importance analysis in the pre-trained video diffusion model and show that LoRAs should be added in the correct layers only to influence the motion and appearance. To further improve the combined results, we introduce a novel test-time training phase when interpolating two LoRAs. Below, we first introduce the preliminaries in the video diffusion model and LoRA as the background knowledge in Sec.~\ref{sec:preliminary}. Then, we give the details of LoRA analysis and the LoRA customization in Sec.~\ref{sec:position-lora} and Sec.~\ref{sec:train-lora}, respectively. Finally, we give the details of the proposed test-time training process after LoRA combinations in Sec.~\ref{sec:ttt}. 

\subsection{Preliminary}
\label{sec:preliminary}
\subsubsection{Text-to-Video Diffusion Models}

Text-to-video diffusion models aim to generate realistic videos from text prompts. Based on the text-to-image latent diffusion model, \ie, stable diffusion model~\cite{LDM}, these methods add the Gaussian noise in the encoded video, \ie latent, in training, where a denoising UNet is used to predict the added step noise. 
Given a text condition $c$, the parameters $\theta$ of the denoising U-Net $\epsilon_\theta$ are optimized using the following reconstruction loss:
\begin{equation}
E_{z_0,c,\epsilon\sim\mathcal{N}\left(0,I\right),t\sim\mathcal{U}\left(0,T\right)}[\left.||\epsilon-\epsilon_\theta\left(z_t,t,\tau_c)\right.||_2^2\right],
\end{equation}
where $z_0$ represents the latent of the encoded training video, $\tau_c$ refers to the pre-trained text encoder $\tau$ with prompt $c$, \(\epsilon\) is the Gaussian noise added to the latent code, and $t$ denotes the time step. The latent variable \(z_t\) at time \(t\) can be expressed as a function of the initial latent variable \(z_0\), the noise \(\epsilon\), and the time step \(t\). Typically, this process is formulated as:
\begin{equation}
   z_t = \sqrt{\alpha_t} z_0 + \sqrt{1 - \alpha_t} \epsilon,
\end{equation}
   where \(\alpha_t\) is a time-dependent scaling factor.
After training, this diffusion model can generate the video from text and noise with a $T$ step progressive sampling process.

\subsubsection{LoRA} Low-Rank Adaptation~(LoRA, \cite{LoRA}) is an effective technique for adapting large pre-trained models to down-streaming tasks with few training-able parameters. To achieve this goal, LoRA introduces a low-rank decomposition-based method to the model's weight matrix, enabling efficient adaptation to new tasks while maintaining the model's original capabilities.

Given the weight matrix ${W}_0\in \mathbb{R}^{m\times n}$ of the original pre-trained model, LoRA uses two low-rank matrices ${A}\in \mathbb{R}^{m \times r}$~($r << m $) and ${B}\in \mathbb{R}^{r\times n}$~($r << n$) to shift the trained distribution according to the new data training. Thanks to the low-rank matrix in ${A}$ and ${B}$, LoRA updates the model more efficiently than the full rank matrix and shows comparable results with full training. Specifically, the new weight matrix $ {W} $ can be represented as: 
\begin{equation}
{W}= {W}_0 + \bigtriangleup {W} = {W}_0+{A}{B}.
\end{equation}

\subsection{Prompt Influence Analysis in VDM}
\label{sec:position-lora}

As introduced previously, directly combining the trained spatial and temporal LoRA results in low-fidelity videos. 
To address this problem, instead of adding LoRAs in each layer, we aim to find \textit{the most crucial layers} for appearance~(or motion) modeling so that we can customize different layers to decouple training and reduce the risk of overfitting.

\begin{figure}[!t]
\centering
\includegraphics[width=0.8\columnwidth]{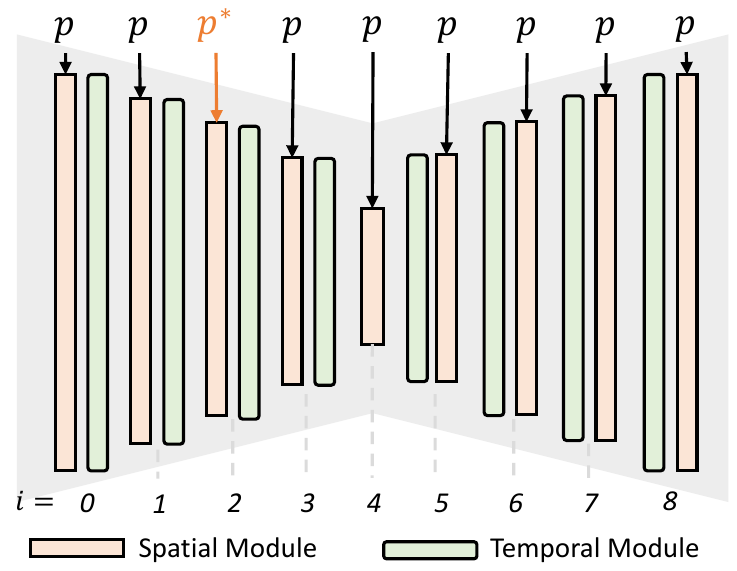}
\caption{Examines the influence of the $i$-th layer on the appearance and motion in video generation. The text prompt 
$p^\ast$ is injected into the $i$-th layer, while the text prompt $p$ is injected into all other layers.}
\label{fig:layer_importance}
\end{figure}

Before introducing our analysis, we first show the overview architecture and some definitions of the current T2V diffusion model to clarify.
As shown in Fig.~\ref{fig:layer_importance}, take AnimateDiff~\cite{Animatediff} as an example, the denoising UNet often comprises four down-sampling modules, one middle module, and four up-sampling modules. Each layer consists of a spatial module and a temporal module, respectively. The spatial module contains several res-blocks~\cite{resnet}, self-attention~\cite{attention} and cross-attention layers while every motion module contains temporal attention layers to model the temporal information. To generate a video, a single text prompt $p$ is embedded into all the cross-attention layers of the pre-trained spatial module. Considering each spatial module as a whole layer, we can mark the video generation process as $V_{p \rightarrow i \in [0,...,8] }$, where $i$ is the index of the layers using $p$. 

Inspired by the customized text-to-image method, \ie, B-LoRA~\cite{B-LoRA}, to find the importance of the prompt embedding in each layer, we analyze by replacing one of the text embeddings in all the pre-trained video diffusion model with the new one and keep other text embeddings unchanged. So, we can observe the relationships between the given hybrid text prompts and the generated video. Formally, considering the original prompt $p$ and the new prompt $p^{*}$, we can generate the video via $V_{p^{*} \rightarrow I, p \rightarrow J}$, where $I$ and $J$ are the module indexes of the prompt $p^{*}$ and $p$, respectively.

\begin{figure}[!t]
\centering
\includegraphics[width=\columnwidth]{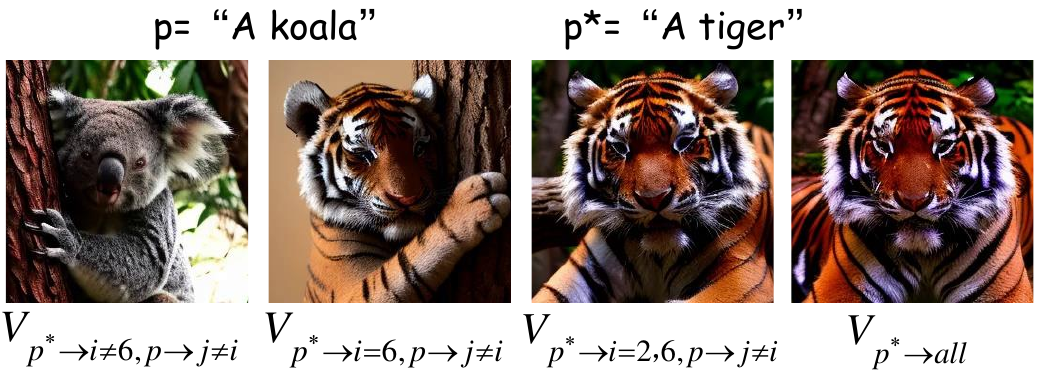}
\vspace{-2em}
\caption{The effect of prompt injection on appearance. Injecting $p^*$ into both $i=2,6$ shows comparable results with $p^*$ all injections. 
% \xiaodong{we may need to show all of the replacement} 
}
\label{fig:appearance_inject}
\vspace{-1em}
\end{figure}

\begin{figure}[!t]
\centering
\includegraphics[width=\columnwidth]{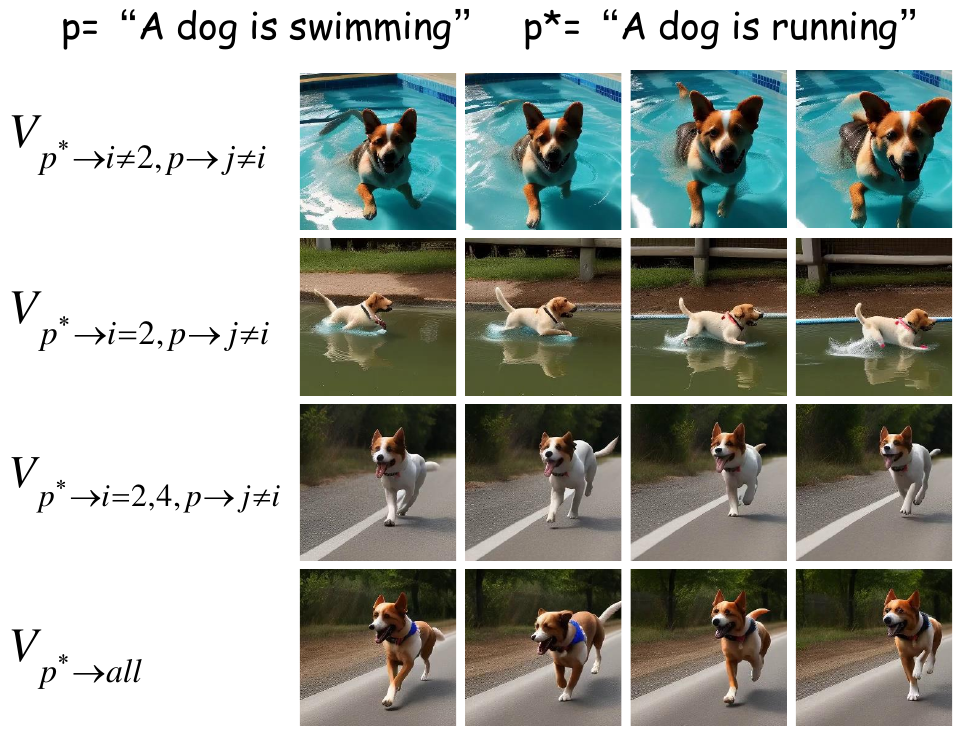}
\vspace{-2em}
\caption{
The effect of prompt injection on motion. Injecting \( p^{*} \) into both \( i=2,4 \) can remove the influence of prompt $p$.}
\label{fig:motion_inject}
\vspace{-1em}
\end{figure}

As shown in Fig.~\ref{fig:appearance_inject}, for appearance, when we only insert $p^{*}$ in the layer 6, $V_{p^{*} \rightarrow i=6, p \rightarrow j\neq i}$, the video predominantly depicts a tiger instead of a koala. However, if we modify the embedding in other locations, the results might not show the correct changes. 
Besides, if we change the prompt in layer $i=6$ only, the tiger's behavior still resembles that of a koala. 
Consequently, we undertake an additional round of the experiment by incorporating a new text embedding replacement within the remaining embeddings. As shown in Fig.~\ref{fig:appearance_inject}, by injecting the new prompt $p^{*}$ at both $i=2,6$, while keeping $p$ unchanged for other layers, the resulting video accurately depicts tiger characteristics without koala interference. This outcome is comparable to injecting ``\textit{a tiger}'' into all layers, indicating that $i=2,6$ are the most important layers for controlling appearance.

Since motion is critical for video, we also conduct a similar experiment replacing the motion-related words in the prompt. \eg, in Fig.~\ref{fig:motion_inject}, we replace one of the text embeddings of the text prompt $p$~(“\textit{a dog is swimming}”) with $p^{*}$~(“\textit{a dog is running}”). Similar to appearance analysis, initially, injecting \( p^\ast \) into a single layer reveals that \( i=2 \) could generate the action described by \( p^\ast \) as shown in the second row of the Fig.~\ref{fig:motion_inject}. However, using only a single \( p^\ast \) injection still results in motion influenced by \( p \) as the dog is running in water, which corresponds to \( p \)'s description of “\textit{a dog is swimming}”. 
This indicates that other layers also significantly impact motion. Building on \( (V_{p^\ast\rightarrow i=2, p\rightarrow j\neq i} )\), we add \( p^\ast \) to other layers and discovered that injecting \( p^\ast \) into \( i=4 \) resulted in a realistic running motion. Therefore, we conclude that the most critical layers for motion are \( i=2 \) and \( i=4 \).

Notice that the behavior of these findings is not only present in the listed figures above. We conduct many experiments to show the effectiveness of our selected layer in the supplementary material.

\begin{figure*}[!t]
\centering
\includegraphics[width=\textwidth]{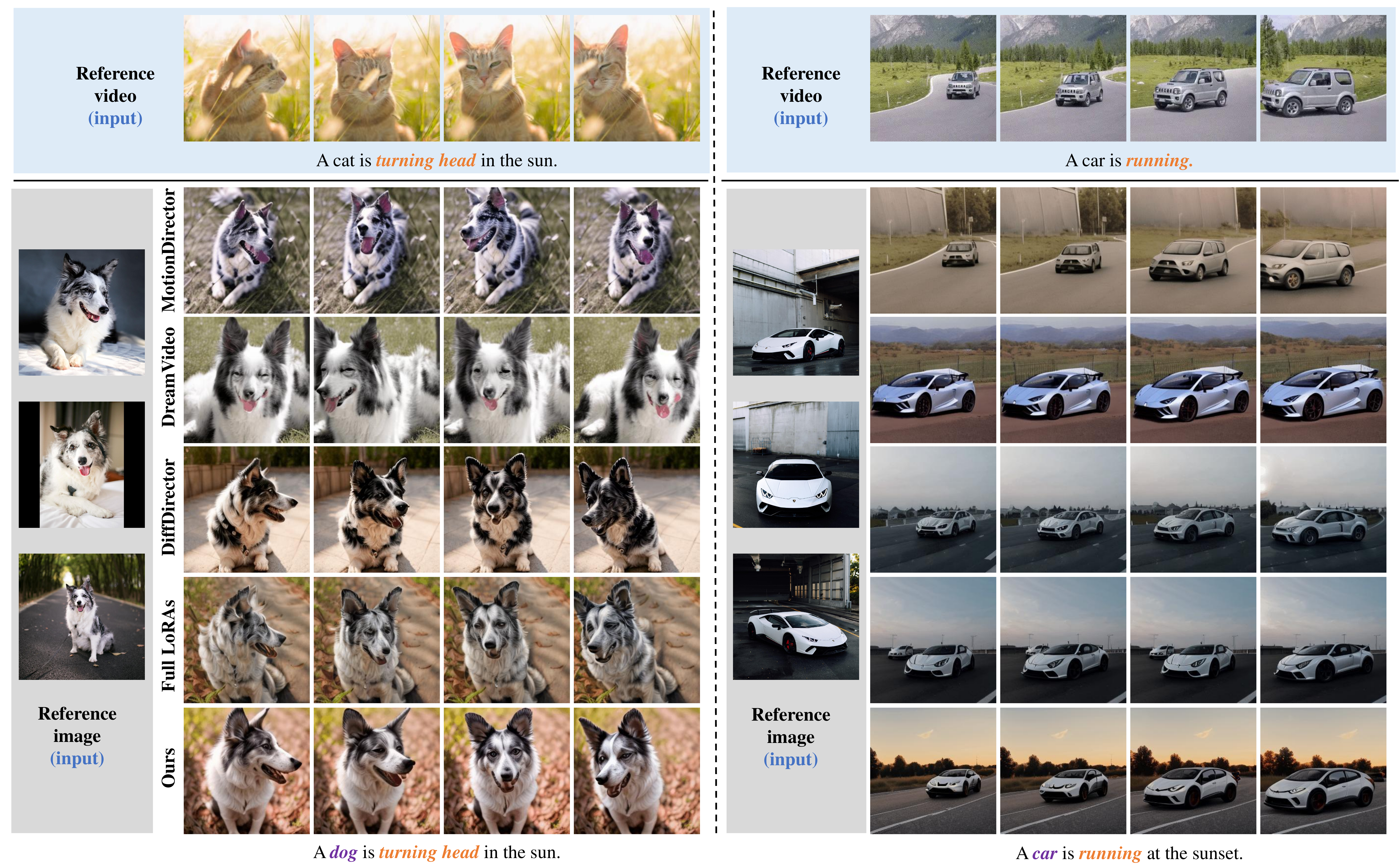}
\vspace{-2em}
\caption{Visual comparison with other state-of-the-art methods.}
\vspace{-1em}
\label{fig:compare_result}
\end{figure*}

\subsection{Train LoRAs Individually for Customization}
\label{sec:train-lora}

As shown in the Fig.~\ref{fig:overall}~(a),~(b), based on the above observations, we train the LoRA weight $ \Delta W_{s}^{2,6} $ on the spatial modules' $2, 6$-th layers for appearance customization.As for the motion customization, current video diffusion models do not have a temporal layer at \(i=4\) and do not have cross-attention at \(i=3\) and \(i=5\), making it impossible to determine their importance through the injection of prompts $p$ and $p^*$. 
Thus, we train the LoRAs $\Delta W^{2,5}_{t}$ on the neighbor temporal layers at $i = 2, 5$ for motion customization. All the experiments are done using the basic training method as introduced in Sec~\ref{sec:preliminary}. Since spatial and temporal LoRAs are placed in different layers, this combination more effectively mitigates conflict in merging weights.

\subsection{Test-time Training for LoRA Combination}
\label{sec:ttt}
Directly combining our individually learned LoRA $\Delta W^{2,5}_{t}$ and $ \Delta W_{s}^{2,6} $ into a single base model already shows better performance than the fully added ones. To further improve the text-video alignment, we propose a novel test-time training process. This is achieved by distilling the knowledge from the single LoRA model as the teacher model. 

As shown in Fig.~\ref{fig:overall}~(c), formally, given the pre-trained T2V diffusion model $\epsilon_{\theta}$, trained LoRA weights of $ \Delta W_{s}^{2,6}$ and $ \Delta W_{t}^{2,5}$ individually, we first inference $f$ step using DDIM~\cite{ddim} to generate a reference latent $z_{f}^{s}$ via $ DDIM(x, \epsilon( c'; \Delta W_{s}^{2,6}) ) $, where $\epsilon( \cdot; \Delta W_{s}^{2,6})$ is the solo appearance customized model and $c'$ are the related prompt to increase the diversity of the reference model. 
% getting the appearance reference latent of $z_{x}$. 
Then, we consider it as the target and add the Gaussian noise $\epsilon'$ to $z_{f}^s$ to obtain $z_{f+1}^{s}$ as the input for finetune. We call it appearance preservation loss $\mathcal{L}_{ap}$ since it measures the difference between the multi-customized model, which can be written as follows:
\begin{equation}
 \mathcal{L}_{ap} = || \epsilon_{\theta}(z_{f+1}^{s}, \tau_c; \Delta \mathcal{W}_{s}^{2,6}, \Delta \mathcal{W}_{t}^{2,5}) - \epsilon' ||^2,
\end{equation}
where $\Delta \mathcal{W}_{s}^{2,6}$ and $\Delta \mathcal{W}_{t}^{2,5}$ are the training-able copy of $ \Delta W_{s}^{2,6}$ and $ \Delta W_{t}^{2,5}$, respectively. 

Meanwhile, we design a temporal preservation loss to have a similar temporal motion as the videos of the original model. Inspired by the temporal debiased loss~\cite{Motiondirector}, for the latent $\epsilon_{i}$ in each frame $i$, we calculate the added noise of each latent as:
\begin{equation}
    \phi(\epsilon_i) = \sqrt{\beta^2 + 1} \epsilon_i - \beta_{anchor},
\end{equation}
where $\beta_{anchor}$ is the selected frame as the anchor frame to remove the influence of the appearance. Similar to appearance finetune, we generate a reference motion latent of $z_{f}^{t}$ via $ DDIM(f, \epsilon( c''; \Delta W_{t}^{2,5}) ) $. We then calculate the temporal preservation loss as follows:
\begin{equation}
 \mathcal{L}_{tp} = || \phi(\epsilon'')  -  \phi(\epsilon_{\theta}(z_{f+1}^{t}, \tau_c; \Delta \mathcal{W}_{s}^{2,6}, \Delta \mathcal{W}_{t}^{2,5})) ||^2, 
\end{equation} where $\epsilon''$ is the added noise for training.

Finally, the motion LoRAs $ \Delta \mathcal{W}_{t}^{2,5}$ and appearance LoRAs $\Delta \mathcal{W}_{s}^{2,6}$ will be jointly trained via optimizing $\mathcal{L}_{ap}$ and $\mathcal{L}_{tp}$, iteratively. After training, they can be directly used for inference.

\begin{table*}[t]
\centering
\caption{Quantitative experimental results for different methods under the numerical evaluation metrics.}
\vspace{-1em}
\begin{tabular}{lcccccccc}
\toprule

\multirow{2}{*}[-1.7ex]{Method} & \multirow{2}{*}[-1.7ex]{${\shortstack{Train-able\\parameters}}$}  & \multicolumn{3}{c}{Objective evaluation} & \multicolumn{4}{c}{User study} \\ 

\cmidrule(r){3-5} \cmidrule(){6-9} 

&  & \makecell[c]{CLIP-T} &\makecell[c]{CLIP-I} &\makecell[c]{Temporal\\consistency} &\makecell[c]{Motion\\similarity} 
&\makecell[c]{Appearance\\similarity} &\makecell[c]{Prompt\\alignment} &\makecell[c]{Video\\quality} \\ \midrule  

Full LoRA & 28.26M &  0.294  & 0.687 & 0.977 & 3.791 & 3.618 & 3.945 & 3.627 \\ 
DreamVideo & 85.00M  &   0.271   & 0.681 & 0.969 & 2.758 & 3.518 & 2.718 & 2.891 \\ 
MotionDirector & 21.26M  & 0.269 & 0.690 & 0.965 & 3.164 & 3.264 & 2.836 & 3.055 \\ 
DiffDirector & 30.46M & 0.287 & 0.685 & 0.971 & 3.773  & 3.491 & 3.727 & 3.427 \\ 
Ours  & \textbf{12.12M} & \textbf{0.301} & \textbf{0.712}  & \textbf{0.978} & \textbf{3.873}  & \textbf{4.045} & \textbf{4.102} & \textbf{3.954} \\ \bottomrule
\end{tabular}
\label{tab:metrics}
\vspace{-0.3cm}
\end{table*}

\section{Experiments}

\subsection{Settings and Implement Details}

\subsubsection{Dataset}
Following previous methods~\cite{customDiffusion, Dreamvideo}, for appearance customization, we collect a total of 13 objects from Dreambooth~\cite{Dreambooth} and CustomDiffusion~\cite{customDiffusion}, including pets, vehicles, toys, \etc. For motion customization, we source 18 sets of videos from the UCF101~\cite{ucf101}, the UCF~\cite{ucfSport} Sports Action datasets, and the DAVIS dataset~\cite{davis}. Each set includes 5 prompts for appearance and motion, covering various scenarios.

\subsubsection{Implementation details}
Our approach uses AnimateDiff~\cite{Animatediff} as the base T2V model, trained with the LION optimizer~\cite{lion}. The learning rate for the spatial LoRA is set at $1e-5$, while the temporal LoRA is trained with a learning rate of $5e-5$. Both types of LoRA are trained for 500 steps with a rank set to 32. In the test-time training phase, we set the learning rate to $1e-6$ and train for 30 steps. During inference, we employ DDIM~\cite{ddim} sampling with 25 steps and a classifier-free guidance~\cite{cfg} scale of 9. We generate 16-frame videos at a resolution of 256×256 and 8 fps. All experiments are performed on a single A6000 GPU.

\subsection{Comparison with other SOTA methods}
To demonstrate the effectiveness of our method in customizing video appearance and motion, we compare it with several current open-source methods. (1)~\textbf{DreamVideo}~\cite{Dreamvideo} achieves appearance and motion customization by separately adding an ID adapter and a Motion Adapter. (2)~\textbf{MotionDirector}~\cite{Motiondirector} is based on the ZeroScope~\cite{Zeroscope} T2V model, with a dual-path LoRAs architecture that decouples the learning of appearance and motion. (3)~\textbf{DiffDirector}~(MotionDirector-AnimateDiff)~\cite{Motiondirector} replaces the original ZeroScope in MotionDirector with the pre-trained AnimateDiff for a fair comparison. (4)~\textbf{Full LoRA}  fine-tunes \textit{all} the spatial and temporal attention blocks using LoRA based on the pre-trained AnimateDiff individually.

\subsubsection{Visual Comparison} As shown in Fig.~\ref{fig:compare_result}, given reference video and images for motion and appearance customization, directly training using full LoRAs shows the visual artifacts. MotionDirector and DiffDirector use the same method with different backbones, they perform well in motion customization. However, they have a very different appearance from the reference images. Besides, DreamVideo seems to overfit the given reference, which has weak connections with the text prompt. Differently, the proposed method achieves a similar behavior to the reference video and also shows high-quality subject customization and text-video alignment.

\begin{figure}[!t]
\centering
\includegraphics[width=\columnwidth]{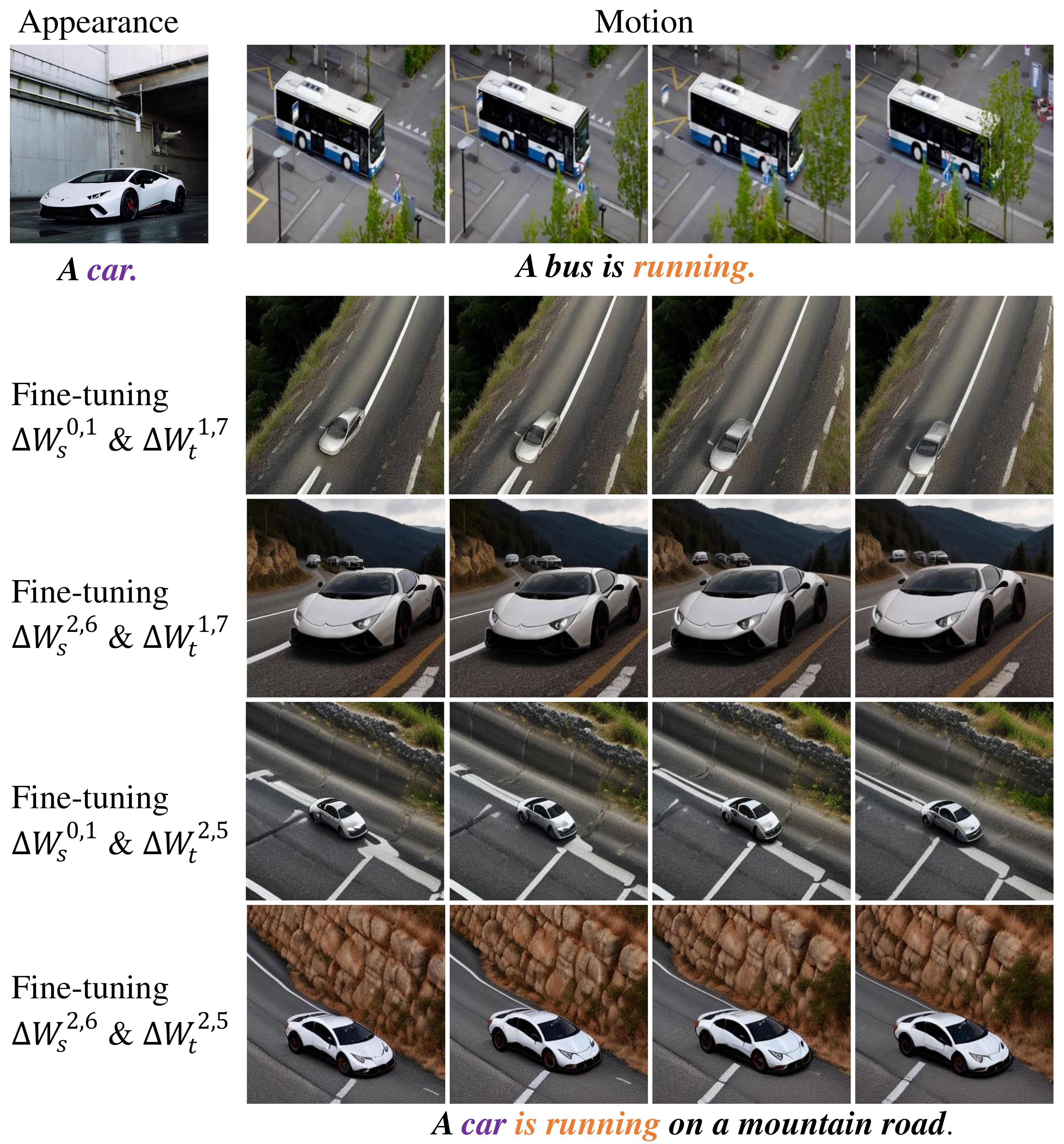}
\vspace{-2em}
\caption{We finetune LoRAs in the specific layers as discussed in Sec.~\ref{sec:position-lora}. The selected layers show the best performance in terms of motion and appearance customization.}
\label{fig:lora_location}
\vspace{-1em}
\end{figure}

\subsubsection{Numerical Comparison}

We employ three metrics to evaluate our method following previous works~\cite{evalcrafter, Dreamvideo}. (1)~\textbf{CLIP-T} measures the alignment between the text and the video by encoding both into embeddings using a clip encoder and calculating their average cosine similarity. (2)~\textbf{CLIP-I} assesses the visual alignment between the generated video and reference images by computing the average similarity of their respective embeddings. (3)~\textbf{Temporal Consistency} evaluates the continuity between consecutive video frames by determining the average cosine similarity of their clip embeddings.
As shown in Tab.~\ref{tab:metrics}, the proposed method shows the best performance on the dataset in terms of visual quality, temporal consistency, and the alignment between the reference image and video. Besides, since we only train the LoRAs on the specific layers, our method has fewer trainingable parameters than others.

% \subsubsection{Visual Comparison}
\subsubsection{User Studies} We also conduct user studies to show the effectiveness of the proposed methods. In detail, we invert 11 participants to vote for 10 generated videos in terms of motion similarity, appearance similarity, prompt alignment, and video quality. Each video will be ranked from different aspects, scoring from 1 to 5. Overall, we get 440 opinions. As shown in Table.~\ref{tab:metrics}, our methods achieve the best score over other state-of-the-art methods.

\subsection{Ablation Studies}

\subsubsection{The effectiveness of training LoRA in crucial layers.}
In Sec.~\ref{sec:position-lora}, we have analyzed the effectiveness of the prompt replacement. Here, we show that training LoRAs on the produced layers improves the performance. As shown in Fig.~\ref{fig:lora_location}, training LoRA in other location, \eg, $ \Delta W_{s}^{0,1}$ and $\Delta W_{t}^{1,7}$ can not customize the appearance and the motion. Fine-tuning the LoRAs on the $ \Delta W_{t}^{2,5}$ or $\Delta W_{s}^{2,6}$ solely performs motion or appearance customization. Ideally, utilizing the $ \Delta W_{s}^{2,6}$ and $\Delta W_{t}^{2,5}$ show the better performance on the multiple customization.

\subsubsection{The effectiveness of reference sampling steps $f$.} 
\begin{figure}[!t]
\centering
\includegraphics[width=\columnwidth]{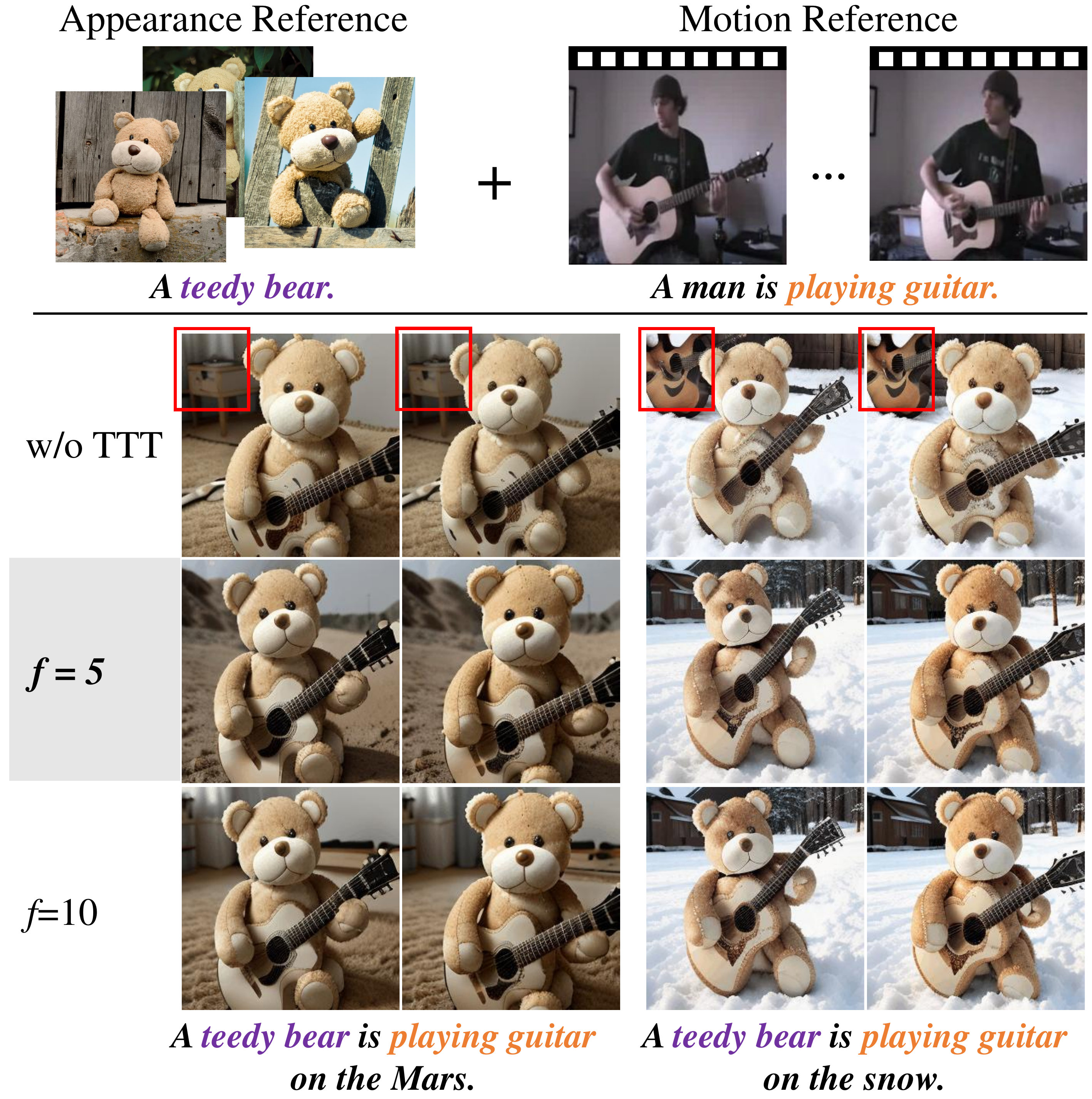}
\vspace{-2em}
\caption{
The influence of the sampling steps of the reference latent in our test-time training stage, using 5 steps of DDIM sampling improves the text-video alignment.
}
\label{fig:ablation_sampling_step}
\vspace{-1em}
\end{figure}

We validate the influence of our reference latent. As shown in Fig.~\ref{fig:ablation_sampling_step}, directly combining the LoRAs w/o TTT shows wrong backgrounds. However, training the model using reference latents will improve the text-video alignment. We evaluate the steps $f$ of the reference latent, where it performs the best at $f = 5$ using DDIM sampling. 

\subsubsection{The effectiveness of test-time training steps.} 
\begin{figure}[!t]
\centering
\includegraphics[width=\columnwidth]{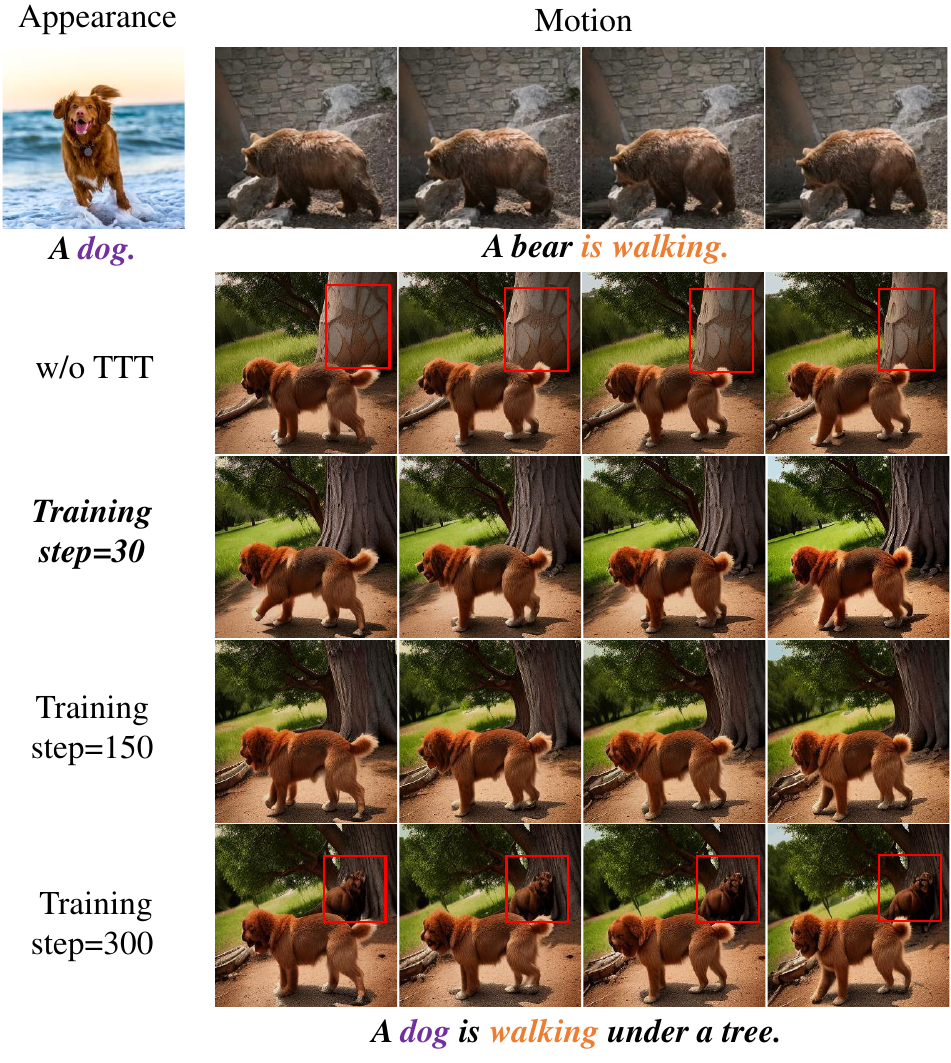}
\vspace{-2em}
\caption{
Ablation study results on test-time training across different numbers of training steps.
}
\vspace{-1em}
\label{fig:ablation_traing_step}
\end{figure}

The number of training steps for TTT significantly affects the results, as shown in Fig.~\ref{fig:ablation_traing_step}. We only require 30 steps to modify a stone pillar in the video to a tree, making it consistent with the prompt description. However, excessive training can lead to artifacts and unnatural effects in the video. 

% \subsection{Limitations} 
% Although our method shows significant improvement with the baseline methods, it still has some limitations. As shown in Fig.~\ref{fig:limitation}, if there exist huge differences between the appearance reference and the motion reference, the results might be influenced. Besides, for small objects, \eg, the golf ball, the current method is limited because of the small inference size. We argue these drawbacks might be solved via the more advanced base models~\cite{videocrafter2, yang2024cogvideox}.  

% \begin{figure}[!t]
% \centering
% \includegraphics[width=\columnwidth]{figs/images/limitation.pdf}
% \vspace{-2em}
% \caption{Limitations. If there is a big disparity between the appearance of the reference and the reference motion, our method may produce unsatisfactory videos.}
% \vspace{-1em}
% \label{fig:limitation}
% \end{figure}
\subsection{Limitations} 
Although our method shows significant improvement with the baseline methods, it still has some limitations. If there exist huge differences between the appearance reference and the motion reference, the results might be influenced. Besides, for small objects, the current method is limited because of the small inference size. We argue these drawbacks might be solved via the more advanced base models~\cite{videocrafter2, yang2024cogvideox}.

\section{Conclusion}
We present a new method to achieve both motion and appearance customization in a single network using LoRAs. To handle the problem of LoRA merges, we first give a detailed analysis to show the most critical layers for video customization in terms of motion and appearance. Then, we design a novel test-time training process that utilizes the pre-trained single LoRA model as guided to further improve the text-video alignment. Based on the proposed methods, our method achieves state-of-the-art appearance and motion customization performance. We believe our findings about the video diffusion model will also benefit the training and the designing of the text-to-video diffusion model.

\section{Acknowledgments}
This work was supported in part by the National Natural Science Foundation of China under Grant 62172067 and Grant 62376046, the Natural Science Foundation of Chongqing for Distinguished Young Scholars under Grant CSTB2022NSCQ-JQX0001, in part by the Natural Science Foundation of Chongqing under Grant CSTB2023NSCQ-MSX0341, the Science and Technology Research Program of Chongqing Municipal Education Commission (Grant No. KJQN202200635)

\bibliography{aaai25}

\end{document}